\documentclass[conference]{IEEEtran}
\IEEEoverridecommandlockouts
\usepackage{algorithm}
\usepackage{algorithmic}
\usepackage{booktabs}
\usepackage{cite}	
\usepackage{comment}
\usepackage{amsmath,amssymb,amsfonts}
\usepackage{cleveref}
\usepackage{algorithmic}
\usepackage{graphicx}
\usepackage{makecell} 
\usepackage{multirow}
\usepackage{textcomp}
\usepackage[table]{xcolor}
\usepackage{xcolor}
\usepackage{tabu}
\usepackage{subfig}
\usepackage{subcaption} 
\usepackage{float}
\def\BibTeX{{\rm B\kern-.05em{\sc i\kern-.025em b}\kern-.08em
    T\kern-.1667em\lower.7ex\hbox{E}\kern-.125emX}}

\usepackage{tikz}
\usetikzlibrary{patterns}   
\usepackage{pgfplots, pgfplotstable}
\usepgfplotslibrary{groupplots}

\begin{document}

\title{Degradation of Feature Space in Continual Learning}


\author{
\IEEEauthorblockN{
    Chiara Lanza\IEEEauthorrefmark{1} \hspace{2cm} 
    Roberto Pereira\IEEEauthorrefmark{1} \hspace{2cm} 
    Marco Miozzo\IEEEauthorrefmark{1} 
}

\IEEEauthorblockN{
     Eduard Angelats\IEEEauthorrefmark{2} \hspace{3cm} Paolo Dini\IEEEauthorrefmark{1}
}

\IEEEauthorblockA{\IEEEauthorrefmark{1}CTTC, Sustainable Artificial Intelligence RU, Castelldefels, Spain \\ 
\IEEEauthorrefmark{2}CTTC, Geomatics RU, Castelldefels, Spain \\ 
\{clanza, rpereira, mmiozzo, eangelats, pdini\}@cttc.es}

}

\maketitle

\begin{abstract}
Centralized training is the standard paradigm in deep learning, enabling models to learn from a unified dataset in a single location. In such setup, isotropic feature distributions naturally arise as a mean to support well-structured and generalizable representations. In contrast, continual learning operates on streaming and non-stationary data, and trains models incrementally, inherently facing the well-known plasticity-stability dilemma. In such settings, learning dynamics tends to yield increasingly anisotropic feature space.

This arises a fundamental question: \textit{should isotropy be enforced to achieve a better balance between stability and plasticity, and thereby mitigate catastrophic forgetting?}

In this paper, we investigate whether promoting feature-space isotropy can enhance representation quality in continual learning. Through experiments using contrastive continual learning techniques on CIFAR-10 and CIFAR-100 data, we find that isotropic regularization fails to improve, and can in fact degrade, model accuracy in continual settings. Our results highlight essential differences in feature geometry between centralized and continual learning, suggesting that isotropy, while beneficial in centralized setups, may not constitute an appropriate inductive bias for non-stationary learning scenarios.

\end{abstract}


\section{Introduction}

Representation learning aims to extract meaningful and structured features from data to support multiple downstream tasks such as classification, recognition, or transfer learning.
In deep learning, well-structured feature spaces tend to encode semantic relationships among data through geometric representations. These relationships are crucial to enable generalization, robustness, and transferability. Task generalization can be promoted by efficient utilization of features \cite{Wang2020UnderstandingCR}.
Contrastive learning \cite{Le-Khac_Healy_Smeaton_2020} has emerged as an effective framework for learning structured feature spaces. 
By pulling semantically similar samples closer and pushing dissimilar samples apart, contrastive objectives yield robust and well-separated representations, both in self-supervised settings (e.g., SimCLR \cite{10.5555/3524938.3525087}) and in supervised variants such as SupCon \cite{10.5555/3495724.3497291}.
It has been shown that well-trained neural networks tend to learn isotropic representations, i.e., features that are uniformly distributed across all directions in the representation space \cite{Guerriero2018DEEPNC}.

While traditional machine learning heavily relies on centralized learning settings where data is aggregated, many real-world applications require learning to happen sequentially due to the streaming nature of the data.
Continual Learning (CL) addresses these more challenging settings where data arrives incrementally in a non-stationary stream.
Nonetheless, a large performance gap between models trained in centralized and continual fashion still exists \cite{Le-Khac_Healy_Smeaton_2020, co2l, dang2024memoryefficientcontinuallearningneural}. 
One of the main challenges is the stability-plasticity dilemma: \textit{models must remain plastic enough to learn new information, and sufficiently stable to preserve previously learned knowledge.}

We hypothesize that to assist generalizations in CL and help manage the stability-plasticity dilemma, features should maintain a well-structured geometric representation throughout the sequential learning process.
In this regard, Contrastive Continual Learning (Co$^2$L) extends supervised contrastive learning to continual learning scenarios \cite{co2l}.
Specifically, Co$^2$L adopts rehearsal strategies to maintain previously learned representations, integrating a replay buffer composed by samples of past data. In addition, it tackles catastrophic forgetting by blending contrastive and knowledge distillation objectives. 

In this work, we compare different contrastive learning approaches by evaluating the isotropy of their feature space and the accuracy performance of the final models.
Additionally, inspired by \cite{rudman2024stableanisotropicregularization}, we study the introduction of an isotropy regularization term in the loss function, aimed at fostering the learning of isotropic feature spaces. 
Different from what one would expect, our numerical experiments have shown that enforcing isotropic feature spaces becomes substantially more difficult in CL settings with respect to centralized. 
This, in turn, degrades both model performance and quality of representations.
We can summarize our main contributions next:
\begin{itemize}   
    \item We propose a mathematical framework simulating data distributions with different levels of isotropy, which assists the interpretation of the isotropy metrics;
    \item We generalize the three-dimensional isotropy metric proposed in \cite{Pandey_2016} for multi-dimensional spaces; 
    \item We compare different contrastive learning techniques in continual settings, considering isotropy levels, accuracy, and geometrical distances among samples in the feature space;
    \item We study the use of an isotropy regularization term in CL and its impact on the learned feature spaces and model accuracy.
\end{itemize}

\section{Background and motivation }

Centralized models tend to learn isotropic feature spaces, where representations are uniformly distributed across feature directions \cite{Guerriero2018DEEPNC}.
Understanding the geometric organization of these learned representations has led to the observation and first definition of the neural collapse phenomenon in  \cite{Papyan_2020} on datasets with multiple classes. Neural collapse is characterized by two main properties: i) representations of all samples from the same class collapse to their class mean vectors; ii) these mean vectors form an equiangular tight frame. This phenomenon has already been studied and applied to enhance transfer learning and adversarial robustness \cite{cisse2017parsevalnetworksimprovingrobustness}. 
As the model approaches convergence during training, the within-class feature variability becomes directionally uniform, and features are evenly distributed around their class mean. These clusters are symmetrically positioned such that their means exhibit maximal angular separation. 
Thus, isotropy provides the local geometry for neural collapse, ensuring that the features of each class are evenly distributed around their mean.

While in centralized settings the geometry of representations converges to a well-defined structure, in CL this geometry changes as new tasks are introduced, and feature distributions become increasingly heterogeneous and anisotropic \cite{goswami2023fecam}. The stability-plasticity dilemma from a feature space perspective shows that gradient projection methods face fundamental challenges in preserving both properties simultaneously \cite{10203314}. In particular, learning new data without access to old samples can damage the learned feature structure and cause the model to forget earlier information \cite{McCloskey_Cohen_1989}.

Neural collapse principles have been proposed to improve continual learning performance in \cite{dang2024memoryefficientcontinuallearningneural}. A neural collapse-inspired feature-classifier alignment for few-shot class-incremental learning has been introduced, called Focal Neural Collapse, utilizing pre-assigned simplex equiangular tight frame prototypes across the entire label space to maintain geometric consistency. Additionally, \cite{montmaur2024neuralcollapse} proposes Neural Continual Collapse, a loss function that directly enforces neural collapse in contrastive continual learning by mapping representations to fixed simplex prototypes, combined with Simplex Structure Distillation to stabilize feature-prototype relationships across tasks without relying heavily on replay buffers. Lastly, \cite{Zhou_Hua_2024} combines anisotropic and isotropic augmentation strategies for continual adversarial defense, suggesting that different geometric properties may be beneficial for different aspects of the learning process. 

Overall, these findings suggest that leveraging geometric properties of feature spaces, such as isotropy and anisotropy, can offer new pathways for improving continual learning performance and robustness.
However, in this work, we argue that enforcing isotropy in continual learning does not inherently lead to more robust or stable representations.
Unlike centralized setups, CL operates under sequential and non-stationary conditions, imposing geometric constraints that limit the emergence of isotropic feature spaces.

\section{Feature Space Geometry}

Class-incremental Continual Learning is a training paradigm where a model receives data in a sequence of experiences $\{e_1, e_2, ... , e_T\}$, each experience $e_t$ containing new information that was not available during previous training phases. The global label space is $\boldsymbol{Y}=\{1,..,C\}$
and each experience $e_t$ contains samples from a subset of classes $\mathbf{y}_t\subset\boldsymbol{Y}$ that have not been seen in previous experiences. The total loss is typically formulated as a sum of per-experience losses $
\mathcal{L}_{\text{tot}} = \sum_{t=1}^T \mathcal{L}_t(e_t)$, where each $\mathcal{L}_t$ balances learning on the current experience $e_t$ with preservation of knowledge from prior experiences.

\begin{figure*}
        \centering
    \includegraphics[width=0.95\linewidth]{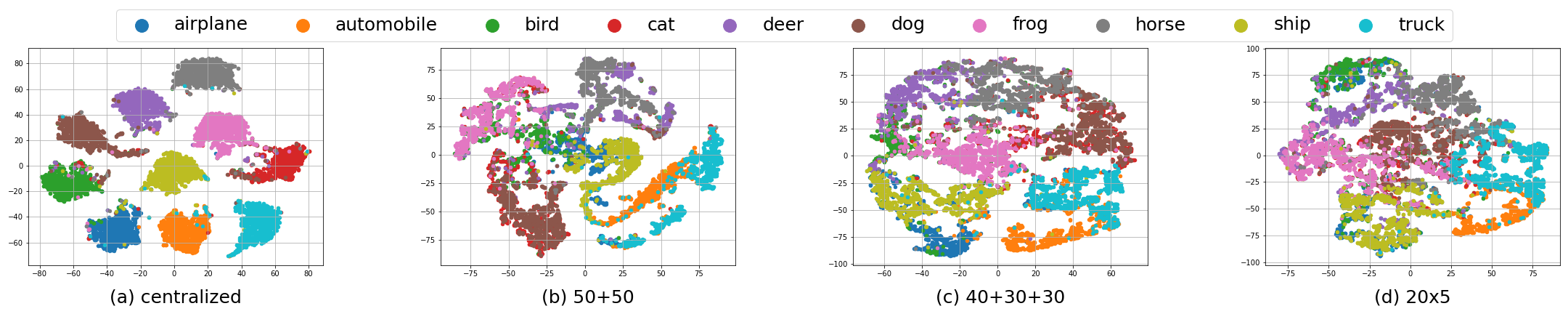}
    \caption{t-SNE visualization for CIFAR-10 dataset with centralized learning and three different CL (CO²L) scenarios: \(50+50\) (2 experiences of 5 classes each)  \(40+30+30\) (3 experiences of 4, 3, and 3 classes), \(20\times5\) (5 experiences of 2 classes each).}
    \label{fig:tsne}
\end{figure*}

In Fig. \ref{fig:tsne}, we illustrate the behavior of the feature space in different learning scenarios (centralized and continual) by reporting the t-SNE of the learned representation with ResNet-18 on CIFAR-10. 
When training is performed in a centralized fashion (Fig. \ref{fig:tsne}a), the feature spaces of the classes result in compact, near-spherical clusters, indicative of isotropic representations. In contrast, when learning occurs continually (Fig. \ref{fig:tsne}b-d), the clusters tend to become elongated and irregular, reflecting a loss of isotropy as the model adapts to new tasks. This phenomenon becomes clearer as the number of experiences increases (from 2 in Fig. \ref{fig:tsne}b, up to 4 in Fig. \ref{fig:tsne}d).

\subsection{Measurements of Isotropy and features geometry}
\label{subsec:metrics}

We consider a set of $K$ multidimensional observations denoted by $\mathbf{X} \in \mathbb{R}^{D \times K}$, where each column $\mathbf{x}_k \in \mathbb{R}^D$ represents a feature vector, e.g., the feature representation of a sample.
A feature space is said to be \emph{isotropic} if it is centered at the origin $\mathbb{E}[\mathbf{X}] = 0$ and its covariance matrix is proportional to the identity matrix:
\begin{align}
\mathbf{\Sigma}
= \mathbb{E}\!\left[(\mathbf{X} - \mathbb{E}[\mathbf{X}])(\mathbf{X} - \mathbb{E}[\mathbf{X}])^\top\right] 
   = \sigma^2 I_D,
\end{align}
where $\sigma^2 > 0$ is a scalar variance term, ${I}_D$ is the $D \times D$ identity matrix, and $\mathbb{E}[\mathbf{X}] \in \mathbb{R}^D$ denotes the expected value of the distribution.

Generally, in a perfect isotropic distribution, its associated covariance matrix has identical eigenvalues, i.e., $\gamma_1 = \gamma_2 = \cdots = \gamma_D = \sigma^2$, where $\boldsymbol{\Gamma} = \{\gamma_i\}_{i=1}^D$ denotes the eigenvalues of $\mathbf{\Sigma}$. In contrast, anisotropy arises when the eigenvalue spectrum is asymmetrical, i.e., when $\lvert \gamma_i - \gamma_j \rvert$ is large for some $i \neq j$.
Consequently, several possible measures of isotropy can be considered, capturing different aspects of variance uniformity in the feature space. Among them, \textit{IsoScore} \cite{rudman-etal-2022-isoscore} quantifies  the deviation of the variance distribution from perfect uniformity and is formally defined as
\begin{equation}
    \text{\textit{IsoScore}}(\mathbf{X}) = 
\frac{
\left(
D - \delta^2(\mathbf{X}) \left(D - \sqrt{D}\right)
\right)^2 - D
}
{D(D - 1)},
\end{equation}
where 
$
\delta(\mathbf{X}) = 
\frac{\lVert  \sqrt{D}(\boldsymbol{\Gamma} / \lVert \boldsymbol{\Gamma} \rVert_2) - I_D \rVert}
{\sqrt{2\left(D - \sqrt{D}\right)}},
$
denotes the isotropy defect. \textit{IsoScore} spans in $[0, 1]$, where $1$ indicates a perfect isotropic distribution.

While \textit{IsoScore} has proven effective in low-dimensional settings \cite{rudman-etal-2022-isoscore}, in high-dimensional spaces its discriminative power is limited. In fact, even a perfectly isotropic Gaussian distribution will lead to small sampling fluctuations in the covariance eigenvalues, which can drastically reduce the \textit{IsoScore}.

Therefore, inspired by \cite{Pandey_2016}, we consider an entropy-based isotropy measure designed for high-dimensional feature distributions, namely \textit{IsoEntropy}.
Let the normalized eigenvalues define a discrete probability distribution over feature-space directions.
The isotropy of this distribution is quantified using the Shannon entropy $
H = - \sum_{j=1}^D  \hat{\gamma}_j \log  \hat{\gamma}_j$, where $\hat{\gamma}_j = {\gamma}_j /\sum_{k=1}^D {\gamma}_k$, for  $j = 1,\dots, D$. Maximum entropy corresponds to the uniform distribution where all eigenvalues are equal, so we normalize it by the maximum possible entropy  for $D$ dimensions, that is $\log D$, resulting  in
\[
\text{\textit{IsoEntropy}}(\mathbf{X}) = \frac{H}{\log D} \in [0, 1].
\]
By construction, $\text{\textit{IsoEntropy}}(\mathbf{X})$ is $1$ if and only if all eigenvalues are equal, corresponding to perfect isotropy. Values approaching zero indicate strong anisotropy, 
i.e., variance concentrated along a small number of principal directions. 
This formulation is particularly suitable for data distributions in high-dimensional spaces, 
where angular similarity measures become less informative, and isotropy is more effectively captured through the relative distribution of variance.

Finally, we also consider distance metrics to evaluate the efficient utilization of geometrical representations in the feature space, i.e., alignment and separability. In particular, we focus on Mahalanobis intra-class and inter-class distances. These metrics correspond, respectively, to the distance between samples within the same class and the distance between the centroids of different classes.
Compared to Euclidean distance, Mahalanobis scales distances by the variance along each direction. As a result, it accounts for anisotropy in the feature distribution and provides a fairer measure of separation when the feature space is not isotropic.
We evaluate the ratio of the average inter-class to intra-class distance: a high ratio means the representations are well separated and compact (good representations), a low ratio means representations are overlapping or scattered.

 \subsection{Synthetic Baselines for Isotropy Measurements}

To better interpret the isotropy metrics, we generate synthetic feature distributions that approximate class clusters with controllable levels of isotropy and anisotropy. We define $M$ Gaussian clusters in $\mathbb{R}^D$, each centered at
 $
 \mu_m = \alpha \, u_m ,
 $
where $u_m$ is a random unit vector and $\alpha$ controls class separability.

Each cluster is sampled from
 \[
 \mathbf{x}^{(m)}_i \sim \mathcal{N}(\mu_m, \Sigma),
\]
with $\Sigma$ defining the variance structure. In the isotropic case, we set $\Sigma = \sigma I$. To introduce anisotropy, we scale the first $k$ principal directions by a factor $(1+\rho)$:
 \[
 \Sigma_{ii} =
 \begin{cases}
 \sigma(1+\rho), & i \le k,\\
 \sigma, & i > k .
 \end{cases}
\]

In our experiments (next section), we vary $\rho$ while keeping $\alpha$ fixed to avoid class-overlap effects. This design choice isolates the intrinsic behavior of the isotropy metrics, enabling a clearer interpretation of the functional dependence on $\rho$. The resulting reference curve serves as a controlling baseline against which the isotropy properties of the feature distributions learned by the models can be compared.

\section{Numerical Evaluation}
\label{sec:results}

\subsection{Learning scenarios}
\label{subsec:settings}

The datasets used in this work are CIFAR-10 and CIFAR-100 \cite{krizhevsky2009learning}.
We train from scratch a ResNet-18 feature extractor and use a 2-layer MLP projection, which maps the model representations into a 128-dimensional latent space \cite{co2l}.
We compare the distribution of this model architecture trained \textbf{centralized} (i.e., with all the classes and data available in a single point) during 1000 epochs,  and in three different continual learning settings, namely:
\begin{itemize}
    \item  \textbf{\(\mathbf{50+50}\)}: the dataset is divided in 2 experiences so that the first experience, trained for 500 epochs, contains 50\% of the classes; the second, trained for 250 epochs, contains the other 50\%.
    \item  \textbf{\(\mathbf{40+30+30}\)}: the dataset is divided in 3 experiences so that the first experience, trained for 500 epochs, contains 40\% of the classes; the second and the third, trained for 150 epochs, contain the other 30\% each.
    \item \textbf{\(\mathbf{20\times5}\)} the dataset is divided into 5 experiences, with 20\% of the classes each. This scenario follows the Co$^2$L epochs scheduling, and has 500 epochs in the first experience and 100 in the others.
\end{itemize}

In the three continual settings, the number of epochs per experience has been chosen to maintain equity among the number of classes per experience. In particular, 500 epochs have been used in the first experience, and a variable number for the subsequent ones, considering 50 epochs per class in the experience.
\subsection{Learning techniques analyzed}
\label{subsec:lt}
The comparison between centralized and continual training scenarios has been performed using 4 different learning techniques:
the state-of-the-art SupCon \cite{10.5555/3495724.3497291}, Co$^2$L \cite{co2l}, and two variants designed ad-hoc in this work, namely SupCP and NCI.

SupCon is used in the asymmetric modified version for Co$^2$L, i.e.,  $\mathcal{L}^{\text{SupCon}}_{\text{asym}} = \sum_{i \in S} - \frac{1}{|P_i|} \sum_{p \in P_i} \log \frac{ \exp( \mathbf{z}_i \cdot \mathbf{z}_p / \tau ) }{ \sum_{k \neq i} \exp( \mathbf{z}_i \cdot \mathbf{z}_k / \tau ) },
$ where $S$ is the set of indexes of current experience samples in the batch, $\tau > 0$ is some temperature hyperparameter and $P_i$ is the index set of positive samples with respect to the anchor $\mathbf{z}_i$. Co$^2$L \cite{co2l} combines the above exposed asymmetric SupCon with a knowledge distillation technique called Instance-wise Relation Distillation (IRD) $, i.e., \mathcal{L}_{\text{IRD}} = \sum_{i=1}^{2N} -p(\tilde{z}_i ; \psi_{\text{past}}) \log p(\tilde{z}_i ; \psi)$, where 
$\mathbf{p}(\cdot)$ is the normalized similarity of a sample to other samples in the batch, and $\psi_{\text{past}}$/$\psi$ are the parameters of the past/current model.

SupCP is a combination of SupCon and SupProto, a prototype-based adaptation of the SupCon, inspired by \cite{dang2024memoryefficientcontinuallearningneural} and \cite{montmaur2024neuralcollapse}, for both centralized and continual settings. SupProto uses class prototypes (i.e., representative centroids $\mathbf{c}_y$ computed as the mean of each class cluster of the feature space) to encourage representations to be closer to their class prototype and farther from , i.e.,  $\mathcal{L}^{\text{SupProto}} = \sum_{i \in I} -\log \frac{ \exp( \mathbf{z}_i \cdot \mathbf{c}_{y_i} / \tau ) }{ \sum_{k \neq y_i} \exp( \mathbf{z}_i \cdot \mathbf{c}_k / \tau ) }$.

Neural Collaps Inspired (NCI) loss inspired by \cite{dang2024memoryefficientcontinuallearningneural} and \cite{montmaur2024neuralcollapse}, adds to the Co$^2$L loss both the SupProto loss and another distillation term (PIRD) with the same approach of IRD in \cite{co2l}, but applied to the prototypes, i.e.,  $\mathcal{L}_{\text{PIRD}} = \sum_{i=1}^{2N} -p(\tilde{\mathbf{z}}_i ; \psi_{\text{past}}, \mathbf{C_{\text{past}}}) \log p(\tilde{\mathbf{z}}_i ; \psi, \mathbf{C_{\text{past}}})$ where $\mathbf{p}(\cdot)$ in this case is the normalized similarity of a sample to all the set of prototypes $C_{\text{past}}$ computed in the previous experience. 
Note that Co$^2$L and NCI are specifically designed for CL and cannot be evaluated in centralized scenarios.

The hyperparameters used those used in \cite{co2l}; moreover the learning rate equal to 0.5, and batch size = 512. 
The distillation power (for IRD and PIRD loss) are $\lambda_{\text{IRD}}, \lambda_{\text{PIRD}}$ equal to 1. The replay buffer dimension used for all learning techniques applied in the CL scenarios is 200 for CIFAR-10 (as in Co$^2$L) and 800 for CIFAR-100 (following CIFAR-10, which has in the last experience 20 samples per class). This buffer has a fixed dimension during the training, and it is updated at each experience so that there is a proportional number of samples for each class already processed in the sequential learning.
Following \cite{co2l}, we perform a linear evaluation by training a simple linear classifier on top of the frozen features extracted by the models for 100 epochs to evaluate the quality of the learned representations. 
\subsection{Isotropy in the CL methods}
\label{subsec:part1}

\begin{figure}[htbp]
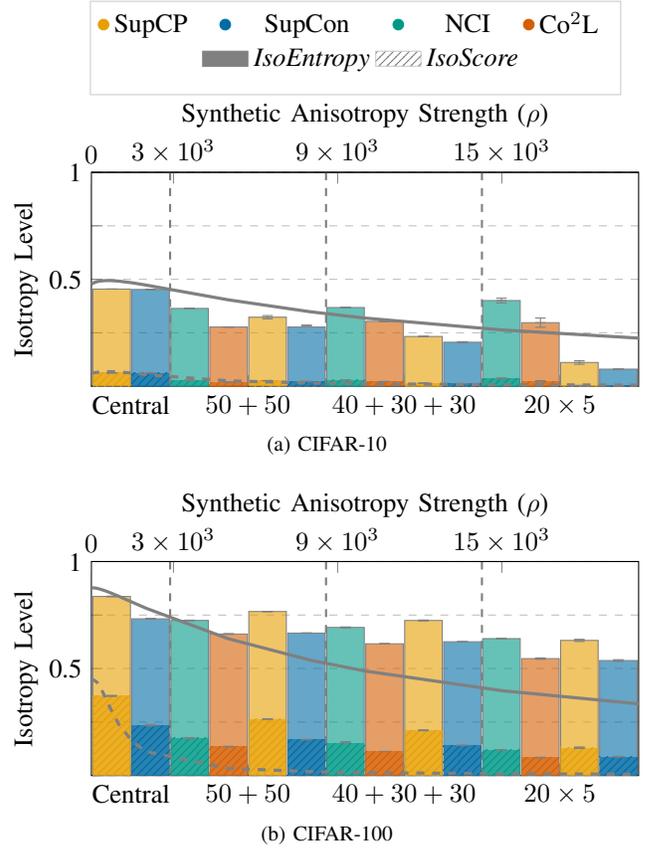

\vspace{-3em}
\centering
\subfloat{\hspace*{-40pt}\definecolor{mycolor1}{rgb}{0.12157,0.46667,0.70588} 
\definecolor{mycolor2}{rgb}{0.17255,0.62745,0.17255} 
\definecolor{mycolor3}{rgb}{1.00000,0.49804,0.05490} 
\definecolor{mycolor4}{rgb}{1.0, 0, 0}


\definecolor{SupCon1}{RGB}{0,107,164}       
\definecolor{CL}{RGB}{213,94,0}    
\definecolor{FNC}{RGB}{0,153,136}           
\definecolor{SupConProto}{RGB}{230,159,0}            



\definecolor{darkgray176}{RGB}{176,176,176}
\definecolor{darkorange25512714}{RGB}{255,127,14}
\definecolor{forestgreen4416044}{RGB}{44,160,44}
\definecolor{lightgray204}{RGB}{204,204,204}
\definecolor{magenta}{RGB}{255,0,255}
\definecolor{orange2551868}{RGB}{255,186,8}
\definecolor{saddlebrown164660}{RGB}{164,66,0}
\definecolor{steelblue31119180}{RGB}{31,119,180}
\definecolor{saddlebrown164660}{RGB}{164,66,0}

\begin{tikzpicture} 
\begin{axis}[
hide axis,
width=0.8\linewidth,
height=1in,
at={(0,0)},
xmin=10,
xmax=50,
ymin=0,
ymax=0.4,
legend style={fill opacity=1,
            draw opacity=1, 
            text opacity=1,
            draw=lightgray204,            
            legend columns=4,
    anchor=north,
    at={(0.7, 0)},
}
]

\addlegendimage{only marks, SupConProto, mark=*, mark size=2pt}
\addlegendentry{SupCP \hspace{1em}}
\addlegendimage{only marks, SupCon1, mark=*, mark size=2pt}

\addlegendentry{SupCon \hspace{1em}};
\addlegendimage{only marks, FNC, mark=*, mark size=2pt}
\addlegendentry{NCI \hspace{1em}};
\addlegendimage{only marks, CL, mark=*, mark size=2pt}
\addlegendentry{Co$^2$L \hspace{1em}};



\addlegendimage{white, opacity=0, only marks}
\addlegendentry{}

\addlegendimage{area legend, color=gray, fill}
\addlegendentry{\textit{IsoEntropy}}

\addlegendimage{area legend, pattern=north east lines, pattern color=gray}
\addlegendentry{\textit{IsoScore}}

\addlegendimage{white, opacity=0, only marks}
\addlegendentry{}
\addlegendimage{white, opacity=0, only marks}
\addlegendentry{}




%
\coordinate (legend) at (axis description cs:0.0,0.0);

\end{axis}

\end{tikzpicture}}
\setcounter{subfigure}{0} 
\subfloat[CIFAR-10]{\input{figures/tikz/100_50_40_20/cifar-10-all}\label{fig:cifar10}}
\\
\subfloat[CIFAR-100]{\input{figures/tikz/100_50_40_20/cifar-100-all}\label{fig:cifar100}}
\caption{
Comparison of \textit{IsoEntropy} and \textit{IsoScore} levels for synthetic distributions and CL methods. For synthetic data, IsoEntropy and IsoScore are depicted using solid and dashed lines, respectively. For CL methods, solid and striped bars represent \textit{IsoEntropy} and \textit{IsoScore}, respectively.
}
\label{fig:curves-bar}
\end{figure}

We start by analyzing the isotropy of the feature space in a centralized scenario and in different CL settings. 
Fig.~\ref{fig:curves-bar} reports the \textit{IsoEntropy} (solid lines and solid-filled bars) and \textit{IsoScore} (dashed lines and striped bars) across these different scenarios.
In Fig. \ref{fig:cifar10} and Fig. \ref{fig:cifar100}, solid and dashed lines depict these two isotropy-based metrics computed on synthetic data 
as a function of the anisotropy strength $\rho$ (top x-axis) with number of clusters $M=10$ and $M=100$, respectively.
The bars, on the other hand, depict the same metrics for models trained with various learning techniques (SupCP, SupCon, NCI, Co$^2$L), with each method represented by a distinct color.

The curves serve as a reference for interpreting how isotropy metrics respond to controlled variations in high-dimensional spaces obtained synthetically. Specifically, increasing $\rho$ leads to more anisotropic spaces, as evidenced by the decline in both \textit{IsoEntropy} (solid lines) and \textit{IsoScore} (dashed lines) values. This trend can also be observed in the results obtained for CL methods. In particular, the maximum isotropy of the synthetic distribution is similar to the isotropy of the centralized scenarios, whereas CL settings present lower isotropy as $\rho$ increases. 
In general, this is confirmed for the different learning methods across both CIFAR-10 and CIFAR-100 datasets. As the training scenarios include more experiences, all methods show a notable reduction in isotropy levels, with the exception of NCI (in green).

Looking at CIFAR-10 (Fig. \ref{fig:cifar10}), traditional learning approaches such as SupCon (yellow) and SupCP (blue), which lack knowledge distillation regularization terms, exhibit substantially lower isotropy in continual settings (e.g., $40+30+30$ and $20\times 5$) compared to their centralized counterparts.
Conversely, methods incorporating knowledge distillation, Co$^2$L (in orange) and NCI (in green), achieve comparable (or even higher for NCI)
isotropy levels with the highest number of experiences ($20\times 5$) than in the other CL scenarios ($50+50$ and $40+30+30$). 
This suggests that catastrophic forgetting mitigation, implemented in Co$^2$L and NCI, contributes to producing more isotropic feature distributions, even under highly non-stationary and streaming settings.

On the other hand, in CIFAR-100 (Fig. \ref{fig:cifar100}), the isotropy of SupCP outperforms the other CL solutions in all the scenarios. We hypothesize that SupProto loss helps in positioning 100 classes' representations more isotropically than with the other learning techniques. 

\begin{table}[h!]
\centering
\setlength{\tabcolsep}{4pt}
\resizebox{0.5\textwidth}{!}{
\begin{tabular}{l l c c c c}
\toprule
Scenario & Method & \multicolumn{2}{c}{CIFAR-10} & \multicolumn{2}{c}{CIFAR-100} \\\cmidrule(lr){3-4} \cmidrule(lr){5-6}
         &        & Accuracy & dist. ratio & Accuracy & dist. ratio \\
\midrule
\multirow{2}{*}{Centralized}
  & SupCP   & 94.93 (0.25) & 2.48 (0.08) & 73.97 (0.20) & 2.35 (0.01) \\
  & SupCon  & 94.98 (0.07) & 10.86 (0.09) & 70.24 (0.32) & 3.32 (0.05) \\
\midrule
\multirow{4}{*}{\(50+50\)}
  & NCI     & 86.94 (0.31) & 1.42 (0.02) & 58.79 (0.21) & 1.45 (0.01) \\
  & Co$^2$L & 84.27 (0.16) & 2.90 (0.02) & 56.37 (0.42) & 1.94 (0.02) \\
  & SupCP   & 78.24 (1.21) & 1.42 (0.02) & 55.89 (0.33) & 1.34 (0.01) \\
  & SupCon  & 75.88 (0.39) & 2.52 (0.04) & 51.65 (0.31) & 1.73 (0.02) \\
\midrule
\multirow{4}{*}{\(40+30+30\)}
  & NCI     & 78.83 (0.76) & 1.24 (0.02) & 50.38 (0.34) & 1.14 (0.01) \\
  & Co$^2$L & 77.22 (0.19) & 2.03 (0.03) & 48.96 (0.33) & 1.67 (0.02) \\
  & SupCP   & 64.80 (0.61) & 1.16 (0.01) & 46.69 (0.39) & 1.01 (0.01) \\
  & SupCon  & 62.95 (0.15) & 1.96 (0.05) & 43.38 (0.60) & 1.35 (0.03) \\
\midrule
\multirow{4}{*}{\(20\times5\)}
  & NCI     & 72.14 (0.58) & 1.06 (0.07) & 44.81 (0.40) & 1.11 (0.01) \\
  & Co$^2$L & 70.64 (1.20) & 1.40 (0.10) & 42.66 (0.65) & 1.55 (0.02) \\
  & SupCP   & 54.47 (3.98) & 1.03 (0.07) & 41.87 (0.42) & 0.99 (0.02) \\
  & SupCon  & 47.58 (1.68) & 1.22 (0.05) & 36.19 (0.48) & 1.27 (0.02) \\
\bottomrule
\end{tabular}
}
\caption{
Accuracy and Mahalanobis-based inter/intra-class distance ratios for CIFAR-10 and CIFAR-100 across centralized and CL scenarios. Values in parentheses indicate standard deviation.
}
\label{tab:iso}
\end{table}

\textbf{Unfortunately, whereas the isotropy tendencies are prominent, they do not consistently translate into higher downstream performance}. Table~\ref{tab:iso} reports classification accuracy together with the Mahalanobis inter/intra ratio in the same scenarios considered in Fig.~\ref{fig:curves-bar}. In the centralized setting, the relationship between isotropy and accuracy is straightforward: on CIFAR-10 SupCon and SupCP have similar isotropy levels and comparable accuracies, and on CIFAR-100 SupCP exhibits higher isotropy and higher accuracy.
However, isotropy and accuracy do not have this behavior in the CL scenarios. 
For example, although NCI obtains similar isotropy levels across different scenarios, its accuracy heavily degrades as the number of experiences increases, dropping from 
$86.94\%$ in the $50 + 50$ scenario 
to $72.14\%$ in the $20\times 5$ one on CIFAR-10 dataset.
Similarly, when comparing NCI and SupCP on the CIFAR-100 dataset, SupCP consistently presents higher isotropy levels than NCI, but its accuracy is lower. 
These results illustrate that, unlike in the centralized setting, higher isotropy in continual learning does not necessarily guarantee higher downstream performance, as confirmed by the results of the distance, presented in what follows.

A similar mismatch also appears when comparing CIFAR-10 and CIFAR-100. Although CIFAR-10 models generally exhibit \textit{lower isotropy levels} than their CIFAR-100 equivalents in the centralized scenarios (reported in Fig.~\ref{fig:curves-bar}a and Fig.~\ref{fig:curves-bar}b, respectively), they achieve \textit{significantly higher accuracies}. For example, in the centralized scenario, SupCon reaches $94.98\%$ on CIFAR-10 and only $70.24\%$ on CIFAR-100, where it has a higher isotropy. 

Regarding the Mahalanobis inter/intra ratio (dist. ratio in the table) reported in Table~\ref{tab:iso}, we expected a correlation between these metrics and the isotropy of the different scenarios. In particular, the inter/intra ratio was expected to increase with the isotropy metrics or the accuracy. However, this trend is not observed in any case. In the centralized scenarios, for example, SupCon shows the highest inter/intra ratio but not the highest \textit{IsoEntropy}, \textit{IsoScore}, or accuracy. A similar lack of correlation can be observed in the CL scenarios, where Co$^2$L consistently achieves the highest ratio, yet this does not correspond to the highest isotropy metrics or accuracy. As the number of experiences increases, the inter/intra ratio distance goes close to 1 in CL settings, indicating partial overlap among several classes. This highlights that dataset-specific factors, such as the number of samples per class, or the learned feature space geometry (distances ratio), can dominate the influence of isotropy when it comes to downstream performance.

These observations suggest that isotropy alone 
is not a sufficient condition to guarantee accuracy performances in continual learning scenarios. High isotropy can be preserved even when geometric distributions of the representation overlap (as in the NCI $20\times 5$ case), and conversely, lower isotropy does not necessarily imply poor performance (as seen in centralized CIFAR-10). 
\subsection{Isotropy as regularization term}
\label{subsec:part2}
Our goal here is to determine whether promoting isotropy through a dedicated regularization term can improve downstream performance in continual learning, or whether it merely shapes the geometry of the feature space without yielding consistent downstream benefits.
To do so, we augment the SupCon loss in the centralized scenario and the Co$^2$L loss in the CL scenarios with a batch-level isotropy regularizer that encourages them to learn more isotropic features. Since the \textit{IsoScore} introduced in Sec.~\ref{subsec:metrics} is not differentiable, we adopt its differentiable equivalent, \textit{IsoScore}$^\ast$ \cite{rudman2024stableanisotropicregularization}, which is stable even when computed on mini-batches. As \textit{IsoScore}$^\ast$ needs to be maximized while the Co$^2$L objective minimized, the combined loss function becomes
$$
\mathcal{L}_{\mathrm{Co}^2\mathrm{L+iso}}=\mathrm{SupCon}+\lambda_{\mathrm{IDR}}\mathrm{IDR}+\lambda_{\mathrm{iso}}(1-\mathrm{\textit{IsoScore}^\ast})
$$
where $\lambda_{\mathrm{IDR}}$ is set to $0$ for the centralized scenario and $1$ in the CL ones. 
This formulation encourages the model not only to learn discriminative and stable representations through contrastive and distillation terms but also to enhance the isotropy of the learned feature space.
When $\lambda_{\mathrm{iso}}$ is set to 0,  the learning method behaves as the standard Co$^2$L approach.

 \begin{table}[!ht]
 \centering
  \resizebox{0.5\textwidth}{!}{
 \begin{tabular}{lc@{\hspace{0.2cm}}c@{\hspace{0.3cm}}c@{\hspace{0.2cm}}c@{\hspace{0.3cm}}c@{\hspace{0.2cm}}c}
 \toprule
 & \multicolumn{2}{c}{\textbf{\(\mathbf{50+50}\)}} 
  & \multicolumn{2}{c}{\textbf{\(\mathbf{40+30+30}\)}} 
 & \multicolumn{2}{c}{\(\mathbf{20\times 5}\)} \\
 \cmidrule(lr){2-3} \cmidrule(lr){4-5} \cmidrule(lr){6-7}  
 \multirow{2}{*}{\(\lambda_{\mathrm{iso}}\)} 
 & IsoScore & Acc 
  & IsoScore & Acc 
 & IsoScore & Acc \\
 & (\(10^{-2}\)) & (\%) 
  & (\(10^{-2}\)) & (\%) 
 & (\(10^{-2}\)) & (\%) \\
 \midrule
 & \multicolumn{6}{c}{\textbf{CIFAR-10}} \\
 \cmidrule(lr){2-7}
 0.0 & 2.30 (0.00) & 84.27 (0.16) 
      &  2.23 (0.02) & 77.22 (0.19)
     & 2.33 (0.32) & 70.64 (1.20) \\
  
 0.5 & 2.03 (0.10) & 81.03 (4.97) 
      &  2.28 (0.02) & 75.52 (0.20)
     & 96.91 (0.36) & 14.74 (1.36) \\
 1.0 & 1.99 (0.02) & 84.33 (0.50) 
      &2.33 (0.11) & 76.34 (0.68)
     & 66.44 (52.81) & 25.90 (16.15) \\
 2.0 & 2.06 (0.03) & 84.19 (0.34) 
     &  2.34 (0.04) & 77.95 (0.54) 
     &  87.46 (1.94) & 31.43 (2.11) \\
 \midrule
 & \multicolumn{6}{c}{\textbf{CIFAR-100}} \\
 \cmidrule(lr){2-7}
 0.0 & 13.53 (0.13) & 56.37 (0.42) 
     & 11.23 (0.07) &48.96 (0.33)
     & 8.38 (0.12) & 42.66 (0.65) \\
0.5 & 14.85 (0.18) & 42.47 (0.90) 
      &11.96 (0.04) &49.14 (0.59)
     & 8.77 (0.15) & 42.33 (0.52) \\
 1.0 & 16.42 (0.38) & 56.66 (0.23) 
      & 12.82 (0.47) & 48.52 (0.43)
     & 9.20 (0.26) & 41.80 (1.21) \\
 2.0 & 19.66 (0.37) & 57.07 (0.42) 
      & 14.37 (0.04) &48.62 (0.55)
     & 9.96 (0.08) & 42.42 (0.75) \\
 \bottomrule
 \end{tabular}
  }
 \caption{Influence of isotropy regularization \(\lambda_{\mathrm{iso}}\) on IsoScore and accuracy under different data distribution scenarios: \(50+50\), \(40+30+30\), and \(20\times 5\). Values in parentheses indicate standard deviation.}
 \label{tab:lambda}
 \end{table}

In the centralized scenario, incorporating the isotropy regularizer yields negligible changes in both accuracy and isotropy metrics, so its results are omitted. This is consistent with our intuition that centralized models trained to convergence already learn highly isotropic representations. As a result, explicitly promoting isotropy offers minimal additional benefit, leaving little room for improving isotropy and accuracy.

We report in Table~\ref{tab:lambda} the resulting accuracy and \textit{IsoScore} levels for the CL scenarios. 
In these scenarios, increasing $\lambda_{\mathrm{iso}}$ does lead to higher \textit{IsoScore} (except for the \(50+50\) CIFAR-10), confirming that the regularizer effectively shapes the geometry of the representation space. 
However, accuracy decreases as the number of experiences increases in all the considered scenarios.
Particularly, for CIFAR-10, this degradation leads to an accuracy below the vanilla SupCon without any distillation (i.e., $\lambda_{\mathrm{IDR}} = 0$), which reaches $47.58,(1.68)$ as previously reported in Table~\ref{tab:iso}.
In CIFAR-100, accuracy also decreases when adding the isotropy regularization term. There are two exceptions, $\lambda_{\mathrm{iso}}=2$ in \(50+50\) and $\lambda_{\mathrm{iso}}=0.5$ in  \(40+30+30\), where the accuracy is slightly higher than without the regularization, but still comparable considering the standard deviations. 
Concluding, differently from what is observed in the centralized setting, enforcing isotropy in CL may conflict with the model’s ability to retain and integrate task-specific information, harming downstream task performance.

\section{Conclusions}
Continual learning is inherently more complex than reproducing centralized approaches, primarily due to the sequential, non-stationary nature of data and the stability-plasticity trade-off. Our analysis shows that isotropy does not guarantee accuracy and effective geometric structure of the representations in CL. Moreover, attempts to enforce isotropy through regularization can adversely affect model performance instead of enhancing it.
Future work could explore alternative geometric properties of the representation space that better reflect the sequential and evolving nature of continual learning. For example, an interesting extension can consider the definition of a regularization term based on Mahalanobis distances able to adapt to the changes of space definition in CL.

\section*{Acknowledgment}

This publication has been funded by the Spanish Project PID2024-160944OB-I00 (SEASIDE) funded by MICIU/AEI/10.13039/501100011033.

{
    \small
    \bibliographystyle{ieeetr}
    \bibliography{references}
}

\end{document}